\documentclass[manuscript, screen]{acmart}
\AtBeginDocument{%
  }

\setcopyright{acmlicensed}
\copyrightyear{2018}
\acmYear{2018}
\acmDOI{XXXXXXX.XXXXXXX}
\acmConference[Integrity 2026]{6th Workshop on Integrity in Social Networks and Media}{August 09--13,
  2026}{Jeju, Korea}
\acmISBN{978-1-4503-XXXX-X/2018/06}

\usepackage[utf8]{inputenc} 
\usepackage[T1]{fontenc}    
\usepackage{hyperref}       
\usepackage{url}            
\usepackage{booktabs}       
\usepackage{amsfonts}       
\usepackage{nicefrac}       
\usepackage{microtype}      
\usepackage{xcolor}         
\usepackage{graphicx} 
\usepackage{lipsum}
\usepackage{float}
\usepackage{wrapfig}
\acmSubmissionID{26}

\begin{document}

\title{Tool-MCoT: Tool Augmented Multimodal Chain-of-Thought for Content Safety Moderation}

\author{Shutong Zhang}
\email{tonyzst@stanford.edu}
\affiliation{%
  \institution{Stanford University}
  \city{Palo Alto}
  \state{California}
  \country{USA}
}

\author{Dylan Zhou}
\author{Yinxiao Liu}
\author{Yang Yang}
\author{Huiwen Luo}
\author{Wenfei Zou}
\affiliation{%
  \institution{Google LLC}
  \city{Mountain View}
  \state{California}
  \country{USA}
}







\renewcommand{\shortauthors}{Zhang et al.}

\begin{abstract}
  The growth of online platforms and user content requires strong content moderation systems that can handle complex inputs from various media types. While large language models (LLMs) are effective, their high computational cost and latency present significant challenges for scalable deployment. To address this, we introduce Tool-MCoT, a small language model (SLM) fine-tuned for content safety moderation leveraging external framework. By training our model on tool-augmented chain-of-thought data generated by LLM, we demonstrate that the SLM can learn to effectively utilize these tools to improve its reasoning and decision-making. Our experiments show that the fine-tuned SLM achieves significant performance gains. Furthermore, we show that the model can learn to use these tools selectively, achieving a balance between moderation accuracy and inference efficiency by calling tools only when necessary.
\end{abstract}



\received{20 February 2007}
\received[revised]{12 March 2009}
\received[accepted]{5 June 2009}

\maketitle

\section{Introduction}
The increasing popularity of online platforms and the explosion of user-generated content has created an urgent need for effective content moderation. 
However, harmful material is often expressed subtly through text, images, or their combinations, making traditional text- or image-only moderation systems insufficient. 
While LLMs are equipped with vision-language understanding and external tool integration can perform nuanced moderation, they are computationally expensive and difficult to deploy at scale. 
In contrast, SLMs are usually faster in terms of generation and require fewer resources to deploy, but they usually have limited ability to understand the complex relationship between modalities.
This motivates a more efficient alternative: a small vision-language model that can reason agentically, leverage external tools to better understand the given image-text pair, then follow a structured chain-of-thought to make moderation decisions. 
In this paper, we present Tool-MCoT, a SLM that leverage external tools to obtain more information then perform further reasoning.
By training these models on synthetic reasoning traces generated by LLMs, we show that the SLM is able to benefit from the tool information and learn to make tool-calling decisions from simple heuristics for content safety moderation via reasoning. 
We summarize our contributions as follows:
\begin{enumerate}
    \item We propose an agentic tool framework consisting of OCR, image captioning, and object detection tools that the model can call to gather more information from the input image.
    \item We fine-tune an SLM that reasons over tool information for content safety moderation. We show that the fine-tuned model achieves performance gains on three open-source datasets.
    \item We further show that the SLM can learn the ability to use tools selectively on harder samples with simple tool calling heuristics, achieving a balance between accuracy and efficiency.
    \vspace{-3mm}
\end{enumerate}

\section{Related Work}
\textbf{Multimodal Content Moderation.} Existing research focusing on multimodal content moderation mainly focuses on classification tasks, aiming to categorize content based on either image or video only~\cite{lu2025kuaimod, qu2024unsafebench}. 
While these methods show promising results in identifying problematic content, most of them operate by combining features from different modalities or employing fusion~\cite{gomez2019mmhs, yuan2024hatefulmemes}, lacking the ability to reason over text-image interactions. 
Our work distinguishes from the previous works by developing methods that enable reasoning across text and image content, moving beyond just the classification to address the reason behind the decisions.

\textbf{Vision Language Model.} Vision-Language Models (VLMs) have demonstrated strong abilities in tasks that involve understanding mixed visual and textual inputs, such as image captioning and visual question answering~\cite{liu2023llava, mohammadkhani2025survey, zhang2024visionlanguagemodelsvisiontasks}.
However, directly applying those general purpose large models for content safety moderation tasks is challenging since they are typically not optimized for this task and specific policies.
As a result, leveraging them for moderation often requires extensive additional prompting.
This results in high computational costs due to the increased input length and complex processing, and large latency due to the size of the model, making real-time or moderation at scale difficult.
In this work, we address these limitations by utilizing a small VLM specifically designed and optimized for the content violation task, which enables cheaper and faster content moderation.

\textbf{Tool Augmented Reasoning.} The integration of tools has significantly improved the reasoning abilities of LLMs, allowing them to access more information and perform complex computations~\cite{lu2025octotools, surismenon2023vipergpt, qin2023toolllmfacilitatinglargelanguage}.
The incorporation of these external tools enables the models to go beyond their internal knowledge and address tasks that require additional information.
However, the benefits of tool augmentation have mainly been observed with large models, as these larger models possess the capability to understand tool functionalities, formulate appropriate queries, and interpret tool outputs in the decision-making process~\cite{wei2023chainofthoughtpromptingelicitsreasoning}.
In contrast, smaller models are typically weaker in reasoning and planning; they often struggle with tool selection, parameterization, and the effective incorporation of external information~\cite{shen2024smallllmsweaktool, yao2023reactsynergizingreasoningacting, li2025small}.
In this work, we will employ a small LLM that is specifically fine-tuned for tool use, thereby overcoming the drawbacks of existing small models.

\section{Methods}
\textbf{Agentic Tool Framework.} The Agentic Tool Framework is a collection of specialized tools designed to augment the capabilities of language models in performing content safety moderation tasks. 
Each tool addresses a specific challenge of multimodal content analysis. The framework contains the following three components:
\begin{enumerate}
    \item \textbf{Optical Character Recognizor (OCR):} This tool extracts text from images, enabling the LLM to process and reason about written content within visual media.
    \item \textbf{Image Captioner:} This tool generates a high-level textual summary of an image, providing the LLM with essential visual context for content moderation.
    \item \textbf{Object Detector:} This tool identifies and locates specific objects within an image, allowing the LLM to recognize potentially harmful objects.
\end{enumerate}
\begin{figure}[!ht]
    \centering
    \includegraphics[width=0.8\linewidth, page=1]{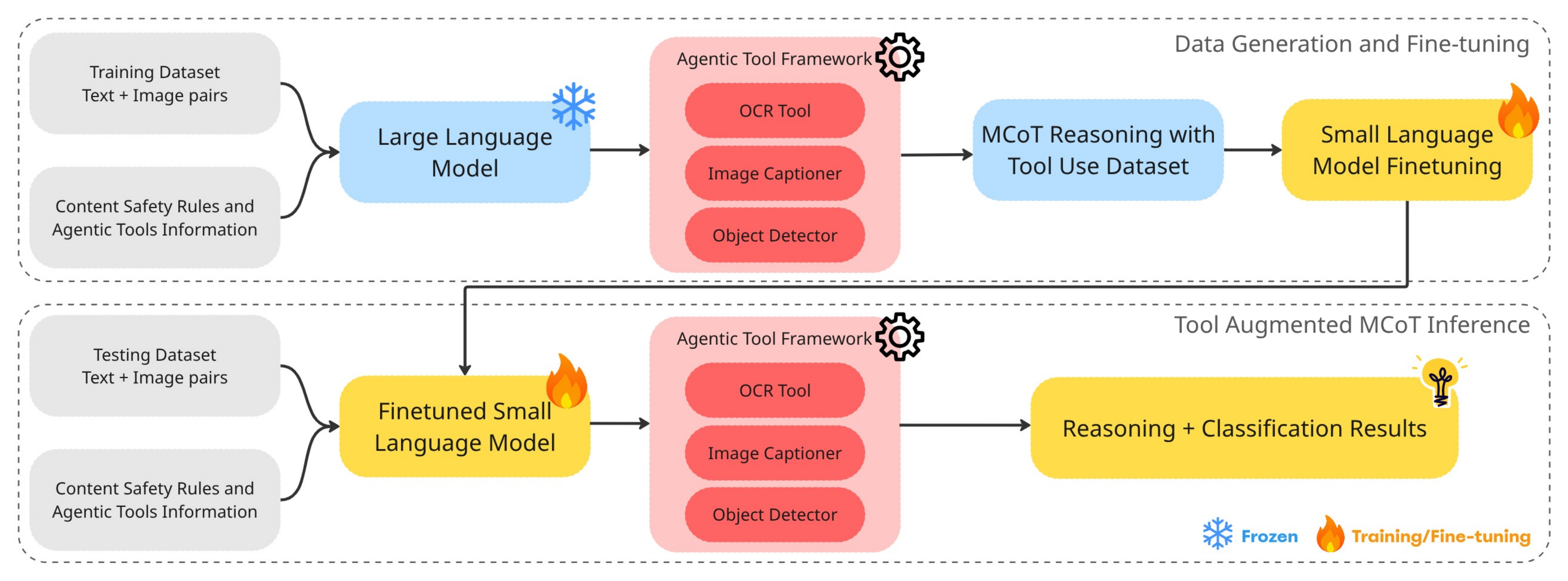}
    \Description{Method figure.}
    \caption{\textbf{An overview of our two-stage pipeline.} We first generate tool reasoning data using a Gemini 2.0 flash as a teacher model, then use the generated reasoning data to fine-tune the SLM. During inference, the fine-tuned SLM utilizes the tool framework to conduct the content safety moderation task.}
    \label{fig:pipeline}
\end{figure}
\vspace{-4mm}
\textbf{MCoT Data Generation.} The first part of the pipeline involves using a LLM to generate Tool MCoT reasoning data to fine-tune the SLM. Specifically, we use the Gemini 2.0 Flash model~\cite{geminiteam2025geminifamilyhighlycapable} as the teacher model to generate the reasoning data. 
We prompt the Gemini model using the input image, text, tool information, and a format requirement to generate reasoning.
Instead of following the STaR introduced in~\cite{zelikman2022starbootstrappingreasoningreasoning}, we prompt the Gemini model to generate ten responses, setting the temperature and top\_p to 1 to increase diversity. 
We then select samples that have at least one reasoning trace with a correct answer for our MCoT reasoning dataset. 
For samples with multiple correct reasoning traces, we randomly select one to use as training data for fine-tuning.
For samples without any correct answer, we then use them as a part of the GRPO training dataset.

\textbf{Small LM finetuning.} The second stage of the pipeline involves using the previously generated MCoT Tool Use dataset to fine-tune a SLM.
In this paper, we use the Gemma3-4B-IT model as the student model~\cite{gemmateam2025gemma3technicalreport}
We employ two fine-tuning approaches: one that enforces tool use for all samples, and another that trains the model to selectively use tools.
In the enforced tool use setting, we force the model to use all tools by including all tool information within the prompt. 
The training process involves two stages: we first conduct LoRA tuning on the reasoning dataset generated by the teacher model~\cite{hu2021loralowrankadaptationlarge}, then further do GRPO using a combined format and answer reward~\cite{shao2024deepseekmathpushinglimitsmathematical}.
In the selective tool use setting, the model learns to selectively call tools following a simple heuristics. 
It is trained to bypass tool use for simple samples that it can predict the correct result without any tool information. For more complex samples, the model is fine-tuned to call the OCR tool for images with overlaid text; call the object detection tool for images with complicated layouts (\textit{i.e.}, has more than five detected objects); and always call image captioning tool. 
The SLM is then fine-tuned on a multi-turn dataset (as shown in Appendix~\ref{sec:tool_calling_conversation}) that follows these rules.

\textbf{Content Moderation.} The final stage involves deploying the fine-tuned SLM for content moderation. During this phase, the SLM interacts with the Agentic Tool Framework to gather insights from multimodal content.

\section{Experiments}
\textbf{Datasets.} We test our proposed method on three multi-modal hate speech datasets.
The MMHS150K dataset contains image-text pairs from Twitter, each manually labeled into one of six classes by three annotators~\cite{gomez2019mmhs}. 
We filtered for samples with a majority label agreement, resulting in 74,177 training and 3,781 testing samples. 
The Hateful Memes dataset is a binary classification dataset contains text overlaid meme, each is hateful or not hateful~\cite{yuan2024hatefulmemes}. 
It contains 8,500 samples in the training set and 1,500 samples in the testing set. 
The UnsafeBench dataset is a content safety multi-class image classification dataset~\cite{qu2024unsafebench}. 
It contains real-world images from the LAION-5B dataset~\cite{schuhmann2022laion5bopenlargescaledataset} and AI-generated images from the Lexica dataset~\cite{shen2024promptstealingattackstexttoimage}. 
The dataset consists of 8,109 training samples and 2,037 testing samples.







\textbf{Fine-tune SLM with enforced tool use.} We first finetuned the small language model using tool reasoning data with LoRA tuning, then further train the model using GRPO. Experiment result shows that the model's performance improves on all three datasets after both LoRA tuning and GRPO.

\begin{table}[!ht]
\centering
\label{tab:gemma3_4b}
\begin{tabular}{l cc cc cc}
\toprule
& \multicolumn{2}{c}{\textbf{HatefulMemes}} & \multicolumn{2}{c}{\textbf{MMHS150K}} &
\multicolumn{2}{c}{\textbf{UnsafeBench}} \\

\cmidrule(lr){2-3} \cmidrule(lr){4-5} \cmidrule(lr){6-7}

\textbf{Model} & \textbf{Accuracy} & \textbf{F1-Score} & \textbf{Accuracy} & \textbf{F1-Score} & \textbf{Accuracy} & \textbf{F1-Score} \\
\midrule

Gemma3-4B-it & 0.646 & 0.613 & 0.395 & 0.458 & 0.486 & 0.523 \\
\textit{w/} LoRA & 0.738 & 0.734 & 0.626 & 0.625 & 0.709 & 0.741 \\
\textit{w/} Reasoning LoRA & 0.777 & 0.763 & 0.655 & \textbf{0.677} & 0.743 & 0.757 \\
\textit{w/} GRPO & \textbf{0.808} & \textbf{0.808} & \textbf{0.662} & 0.674 & \textbf{0.767} & \textbf{0.773} \\

\bottomrule
\end{tabular}
\vspace{2mm}
\caption{\textbf{Fine-tune small language model using tool reasoning dataset.} }
\end{table}
\vspace{-5mm}
\textbf{Fine-tune SLM to selectively use tool.} We further investigate whether the small language model is able to learn when and what tools to use. 
We finetune the model using tool conversation data then compare it with the base model and the tool reasoning LoRA-tuned model (Force Tool) on the MMHS150K dataset. 
Experiment results show that after fine-tuning the model to actively select tools, the tool calling ratio and run time (tested on a single Nvidia A100 GPU) reduced significantly while the performance remain almost unchanged.

\begin{table}[!ht]
\centering
\label{tab:choose_tool}
\begin{tabular}{l cccccc}
\toprule

\textbf{Model} & \textbf{Accuracy} & \textbf{F1-Score} & \textbf{OCR Ratio} & \textbf{Captioner Ratio} & \textbf{Detector Ratio} & \textbf{Run Time (s)} \\
\midrule

Gemma3-4B-it & 0.395 & 0.458 & - & - & - & 0.13 \\
\textit{w/} Force Tool & \textbf{0.655} & \textbf{0.677} & 1 & 1 & 1 & 1.67 \\
\textit{w/} Select Tool & 0.634 & 0.651 & \textbf{0.203} & \textbf{0.484} & \textbf{0.263} & 0.86\\

\bottomrule
\end{tabular}
\vspace{2mm}
\caption{\textbf{Finetune small language model to actively call tools when necessary.} }
\end{table}
\vspace{-7mm}

\section{Conclusion}
In this paper, we presented Tool-MCoT, an efficient and effective method for multimodal content moderation by augmenting a small language model with external tools. 
We demonstrate that a fine-tuned SLM can learn to leverage tool information to improve its reasoning and classification performance. We also show that the SLM can even learn to selectively use tools from simple tool calling heuristics to achieve a balance between performance and efficiency.
By employing a pipeline that uses a large language model as a teacher to generate synthetic, tool-augmented reasoning data, we successfully transferred advanced reasoning capabilities to a much smaller, more resource-efficient model. 
Future work could explore methods such as using prompt optimization~\cite{pryzant2023automaticpromptoptimizationgradient} to generate better prompts for the model, jointly training the language model's reasoning head with a classification head to improve inference efficiency, or investigating more advanced methodologies to enhance the SLM's selective tool-use capabilities beyond simple heuristics.

\bibliographystyle{ACM-Reference-Format}
\bibliography{ref}

@misc{lu2025kuaimod,
      title={VLM as Policy: Common-Law Content Moderation Framework for Short Video Platform}, 
      author={Xingyu Lu and Tianke Zhang and Chang Meng and Xiaobei Wang and Jinpeng Wang and YiFan Zhang and Shisong Tang and Changyi Liu and Haojie Ding and Kaiyu Jiang and Kaiyu Tang and Bin Wen and Hai-Tao Zheng and Fan Yang and Tingting Gao and Di Zhang and Kun Gai},
      year={2025},
      eprint={2504.14904},
      archivePrefix={arXiv},
      primaryClass={cs.SI},
      url={https://arxiv.org/abs/2504.14904}, 
}

@misc{qu2024unsafebench,
      title={UnsafeBench: Benchmarking Image Safety Classifiers on Real-World and AI-Generated Images}, 
      author={Yiting Qu and Xinyue Shen and Yixin Wu and Michael Backes and Savvas Zannettou and Yang Zhang},
      year={2024},
      eprint={2405.03486},
      archivePrefix={arXiv},
      primaryClass={cs.CR}
}

@misc{gomez2019mmhs,
      title={Exploring Hate Speech Detection in Multimodal Publications}, 
      author={Raul Gomez and Jaume Gibert and Lluis Gomez and Dimosthenis Karatzas},
      year={2019},
      eprint={1910.03814},
      archivePrefix={arXiv},
      primaryClass={cs.CV},
      url={https://arxiv.org/abs/1910.03814}, 
}

@INPROCEEDINGS {yuan2024hatefulmemes,
    author = { Yuan, Jialin and Yu, Ye and Mittal, Gaurav and Hall, Matthew and Sajeev, Sandra and Chen, Mei },
    booktitle = { 2024 IEEE/CVF Winter Conference on Applications of Computer Vision (WACV) },
    title = {{ Rethinking Multimodal Content Moderation from an Asymmetric Angle with Mixed-modality }},
    year = {2024},
    month =Jan
}

@misc{liu2023llava,
  title={Visual Instruction Tuning}, 
  author={Liu, Haotian and Li, Chunyuan and Wu, Qingyang and Lee, Yong Jae},
  publisher={NeurIPS},
  year={2023},
}

@article{mohammadkhani2025survey,
  title={A Survey on Bridging VLMs and Synthetic Data},
  author={Mohammadkhani, Mohammad Ghiasvand and Momtazi, Saeedeh and Beigy, Hamid},
  journal={Authorea Preprints},
  year={2025},
  publisher={Authorea}
}

@article{lu2025octotools,
    title={OctoTools: An Agentic Framework with Extensible Tools for Complex Reasoning},
    author={Lu, Pan and Chen, Bowen and Liu, Sheng and Thapa, Rahul and Boen, Joseph and Zou, James},
    journal = {arXiv preprint arXiv:2502.11271},
    year={2025}
}

@article{surismenon2023vipergpt,
    title={ViperGPT: Visual Inference via Python Execution for Reasoning},
    author={D\'idac Sur\'is and Sachit Menon and Carl Vondrick},
    journal={Proceedings of IEEE International Conference on Computer Vision (ICCV)},
    year={2023}
}

@misc{qin2023toolllmfacilitatinglargelanguage,
      title={ToolLLM: Facilitating Large Language Models to Master 16000+ Real-world APIs}, 
      author={Yujia Qin and Shihao Liang and Yining Ye and Kunlun Zhu and Lan Yan and Yaxi Lu and Yankai Lin and Xin Cong and Xiangru Tang and Bill Qian and Sihan Zhao and Lauren Hong and Runchu Tian and Ruobing Xie and Jie Zhou and Mark Gerstein and Dahai Li and Zhiyuan Liu and Maosong Sun},
      year={2023},
      eprint={2307.16789},
      archivePrefix={arXiv},
      primaryClass={cs.AI},
      url={https://arxiv.org/abs/2307.16789}, 
}

@misc{shen2024smallllmsweaktool,
      title={Small LLMs Are Weak Tool Learners: A Multi-LLM Agent}, 
      author={Weizhou Shen and Chenliang Li and Hongzhan Chen and Ming Yan and Xiaojun Quan and Hehong Chen and Ji Zhang and Fei Huang},
      year={2024},
      eprint={2401.07324},
      archivePrefix={arXiv},
      primaryClass={cs.AI},
      url={https://arxiv.org/abs/2401.07324}, 
}

@misc{zhang2024visionlanguagemodelsvisiontasks,
      title={Vision-Language Models for Vision Tasks: A Survey}, 
      author={Jingyi Zhang and Jiaxing Huang and Sheng Jin and Shijian Lu},
      year={2024},
      eprint={2304.00685},
      archivePrefix={arXiv},
      primaryClass={cs.CV},
      url={https://arxiv.org/abs/2304.00685}, 
}

@misc{yao2023reactsynergizingreasoningacting,
      title={ReAct: Synergizing Reasoning and Acting in Language Models}, 
      author={Shunyu Yao and Jeffrey Zhao and Dian Yu and Nan Du and Izhak Shafran and Karthik Narasimhan and Yuan Cao},
      year={2023},
      eprint={2210.03629},
      archivePrefix={arXiv},
      primaryClass={cs.CL},
      url={https://arxiv.org/abs/2210.03629}, 
}

@article{li2025small,
  title={Small Models Struggle to Learn from Strong Reasoners},
  author={Li, Yuetai and Yue, Xiang and Xu, Zhangchen and Jiang, Fengqing and Niu, Luyao and Lin, Bill Yuchen and Ramasubramanian, Bhaskar and Poovendran, Radha},
  journal={arXiv preprint arXiv:2502.12143},
  year={2025}
}

@misc{wei2023chainofthoughtpromptingelicitsreasoning,
      title={Chain-of-Thought Prompting Elicits Reasoning in Large Language Models}, 
      author={Jason Wei and Xuezhi Wang and Dale Schuurmans and Maarten Bosma and Brian Ichter and Fei Xia and Ed Chi and Quoc Le and Denny Zhou},
      year={2023},
      eprint={2201.11903},
      archivePrefix={arXiv},
      primaryClass={cs.CL},
      url={https://arxiv.org/abs/2201.11903}, 
}

@misc{geminiteam2025geminifamilyhighlycapable,
      title={Gemini: A Family of Highly Capable Multimodal Models}, 
      author={Gemini Team},
      year={2025},
      eprint={2312.11805},
      archivePrefix={arXiv},
      primaryClass={cs.CL},
      url={https://arxiv.org/abs/2312.11805}, 
}

@misc{gemmateam2025gemma3technicalreport,
      title={Gemma 3 Technical Report}, 
      author={Gemma Team},
      year={2025},
      eprint={2503.19786},
      archivePrefix={arXiv},
      primaryClass={cs.CL},
      url={https://arxiv.org/abs/2503.19786}, 
}

@misc{zelikman2022starbootstrappingreasoningreasoning,
      title={STaR: Bootstrapping Reasoning With Reasoning}, 
      author={Eric Zelikman and Yuhuai Wu and Jesse Mu and Noah D. Goodman},
      year={2022},
      eprint={2203.14465},
      archivePrefix={arXiv},
      primaryClass={cs.LG},
      url={https://arxiv.org/abs/2203.14465}, 
}

@misc{hu2021loralowrankadaptationlarge,
      title={LoRA: Low-Rank Adaptation of Large Language Models}, 
      author={Edward J. Hu and Yelong Shen and Phillip Wallis and Zeyuan Allen-Zhu and Yuanzhi Li and Shean Wang and Lu Wang and Weizhu Chen},
      year={2021},
      eprint={2106.09685},
      archivePrefix={arXiv},
      primaryClass={cs.CL},
      url={https://arxiv.org/abs/2106.09685}, 
}

@misc{shao2024deepseekmathpushinglimitsmathematical,
      title={DeepSeekMath: Pushing the Limits of Mathematical Reasoning in Open Language Models}, 
      author={Zhihong Shao and Peiyi Wang and Qihao Zhu and Runxin Xu and Junxiao Song and Xiao Bi and Haowei Zhang and Mingchuan Zhang and Y. K. Li and Y. Wu and Daya Guo},
      year={2024},
      eprint={2402.03300},
      archivePrefix={arXiv},
      primaryClass={cs.CL},
      url={https://arxiv.org/abs/2402.03300}, 
}

@misc{schuhmann2022laion5bopenlargescaledataset,
      title={LAION-5B: An open large-scale dataset for training next generation image-text models}, 
      author={Christoph Schuhmann and Romain Beaumont and Richard Vencu and Cade Gordon and Ross Wightman and Mehdi Cherti and Theo Coombes and Aarush Katta and Clayton Mullis and Mitchell Wortsman and Patrick Schramowski and Srivatsa Kundurthy and Katherine Crowson and Ludwig Schmidt and Robert Kaczmarczyk and Jenia Jitsev},
      year={2022},
      eprint={2210.08402},
      archivePrefix={arXiv},
      primaryClass={cs.CV},
      url={https://arxiv.org/abs/2210.08402}, 
}

@misc{shen2024promptstealingattackstexttoimage,
      title={Prompt Stealing Attacks Against Text-to-Image Generation Models}, 
      author={Xinyue Shen and Yiting Qu and Michael Backes and Yang Zhang},
      year={2024},
      eprint={2302.09923},
      archivePrefix={arXiv},
      primaryClass={cs.CR},
      url={https://arxiv.org/abs/2302.09923}, 
}

@misc{loshchilov2019decoupledweightdecayregularization,
      title={Decoupled Weight Decay Regularization}, 
      author={Ilya Loshchilov and Frank Hutter},
      year={2019},
      eprint={1711.05101},
      archivePrefix={arXiv},
      primaryClass={cs.LG},
      url={https://arxiv.org/abs/1711.05101}, 
}

@misc{pryzant2023automaticpromptoptimizationgradient,
      title={Automatic Prompt Optimization with "Gradient Descent" and Beam Search}, 
      author={Reid Pryzant and Dan Iter and Jerry Li and Yin Tat Lee and Chenguang Zhu and Michael Zeng},
      year={2023},
      eprint={2305.03495},
      archivePrefix={arXiv},
      primaryClass={cs.CL},
      url={https://arxiv.org/abs/2305.03495}, 
}

\appendix

\section{Training Setup}
\subsection{Hardware Selection}
All experiments were conducted on a cluster of eight NVIDIA H100 (80GB) GPUs. Specifically, LoRA fine-tuning was performed using Distributed Data Parallel (DDP), while the GRPO training utilized DeepSpeed ZeRO Stage 3 to manage the increased memory overhead of group-based reinforcement learning.

\subsection{LoRA Tuning Setup}
We use the following hyperparameters for the LoRA tuning:
\begin{itemize}
    \item \textbf{Optimizer}: AdamW~\cite{loshchilov2019decoupledweightdecayregularization} with cosine scheduler with 0.1 weight decay and warmup ratio
    \item \textbf{Data Precision}: torch.bfloat16
    \item \textbf{Learning Rate}: 1e-5
    \item \textbf{Epochs}: 2 for the MMHS150K Dataset, 3 for the HarefulMemes and UnsafeBench Dataset
    \item \textbf{Per GPU Batch Size}: 1
    \item \textbf{Gradient Accumulation Step}: 16
    \item \textbf{LoRA Module}: q\_proj, v\_proj, k\_proj, o\_proj, gate\_proj, up\_proj, down\_proj
    \item \textbf{LoRA Rank}: 16
    \item \textbf{LoRA Alpha}: 32
    \item \textbf{LoRA Dropout}: 0.1
\end{itemize}

\subsection{GRPO Setup}
We use the following hyperparameters for the GRPO tuning:
\begin{itemize}
    \item \textbf{Optimizer}: AdamW~\cite{loshchilov2019decoupledweightdecayregularization} with cosine scheduler with 0.1 weight decay and warmup ratio
    \item \textbf{Data Precision}: torch.bfloat16
    \item \textbf{Learning Rate}: 1e-5
    \item \textbf{Epochs}: 1e-6
    \item \textbf{Per GPU Batch Size}: 1
    \item \textbf{Gradient Accumulation Step}: 8
    \item \textbf{Loss Type}: dr\_grpo
    \item \textbf{Num Iterations}: 2
    \item \textbf{Num Generations}: 16
    \item \textbf{Reward Type}: [answer reward, format reward]
    \item \textbf{Reward Weight}: [4, 1]
    \item \textbf{LoRA Module}: q\_proj, v\_proj, k\_proj, o\_proj, gate\_proj, up\_proj, down\_proj
    \item \textbf{LoRA Rank}: 16
    \item \textbf{LoRA Alpha}: 32
    \item \textbf{LoRA Dropout}: 0.1
\end{itemize}

\subsection{Inference Setup}
We do inference using the following hyperparameters:
\begin{itemize}
    \item \textbf{do\_sample}: True
    \item \textbf{temperature}: 0.1
    \item \textbf{top\_p}: 0.3
    \item \textbf{new\_max\_token}: 512
\end{itemize}

\section{Qualitative result}
The qualitative examples shown in the image below demonstrate how tool information and internal reasoning improves model accuracy on the Hateful Memes dataset by enabling complex contextual grounding. Without reasoning, the model often relies on surface-level heuristics, leading to incorrect classifications of reclaimed slurs in positive contexts or failing to identify hateful memes that require historical knowledge, such as recognizing a Nazi crematorium. By utilizing information provided by the internal tool and the \textbf{<think>} process, it identifies the athletic setting in the first image and the historical significance of the crematorium in the second, allowing it to catch the nuance that a standard model misses.

\begin{figure}[!ht]
    \centering
    \includegraphics[width=1.0\linewidth]{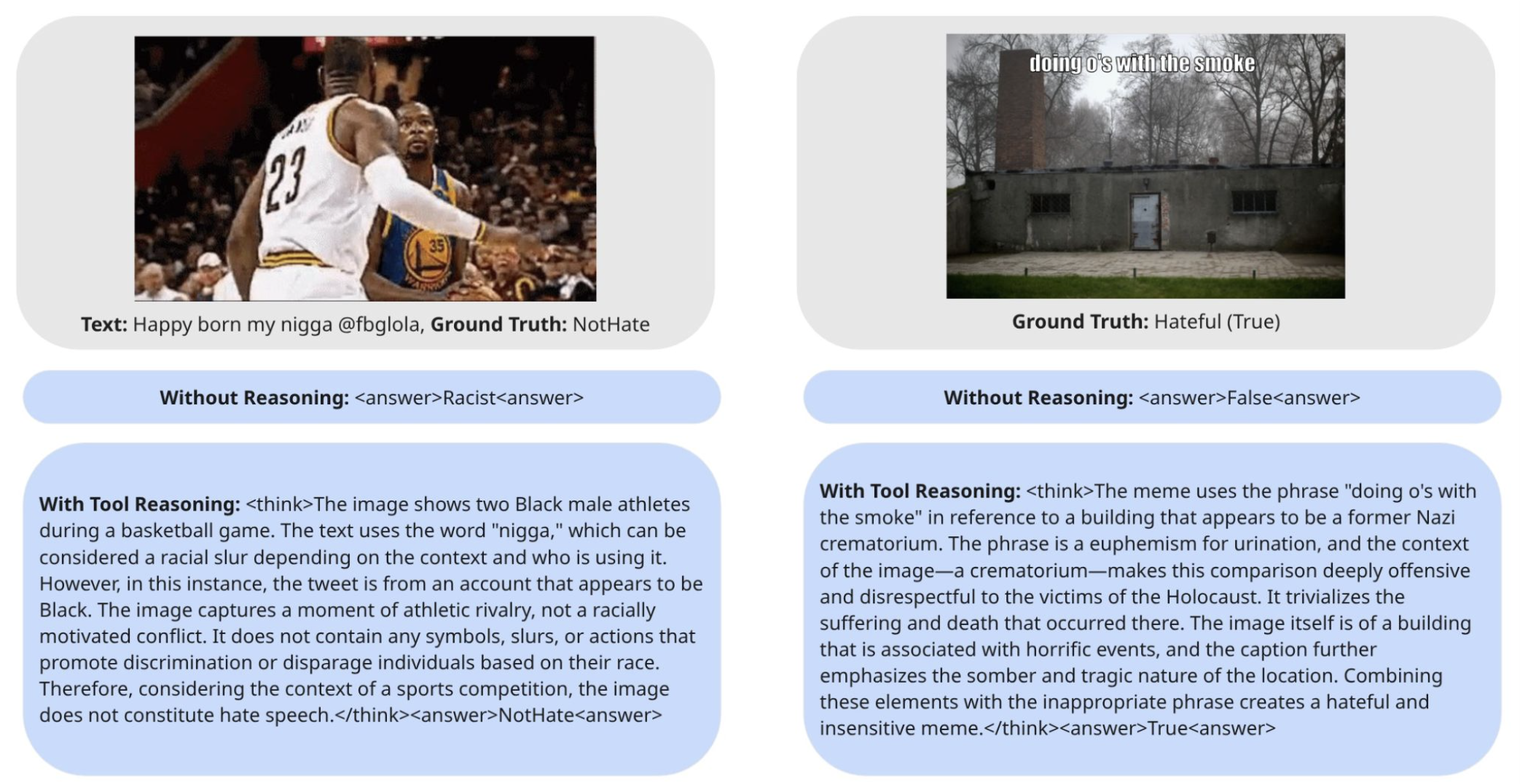}
    \vspace{-3mm}
    \Description{Qualitative comparison showing two examples of model outputs with and without tool reasoning. The tool reasoning model identifies the intended context more accurately than the standard model.}
    \caption{\textbf{Qualitative comparison between model outputs with and without tool reasoning.} The tool reasoning model (bottom) correctly identifies the nuanced context in both cases, where the standard model fails.}
    \label{fig:pipeline}
\end{figure}

\section{Tool Calling Conversation}
\label{sec:tool_calling_conversation}
\begin{figure}[!ht]
    \centering
    \includegraphics[width=0.5\linewidth]{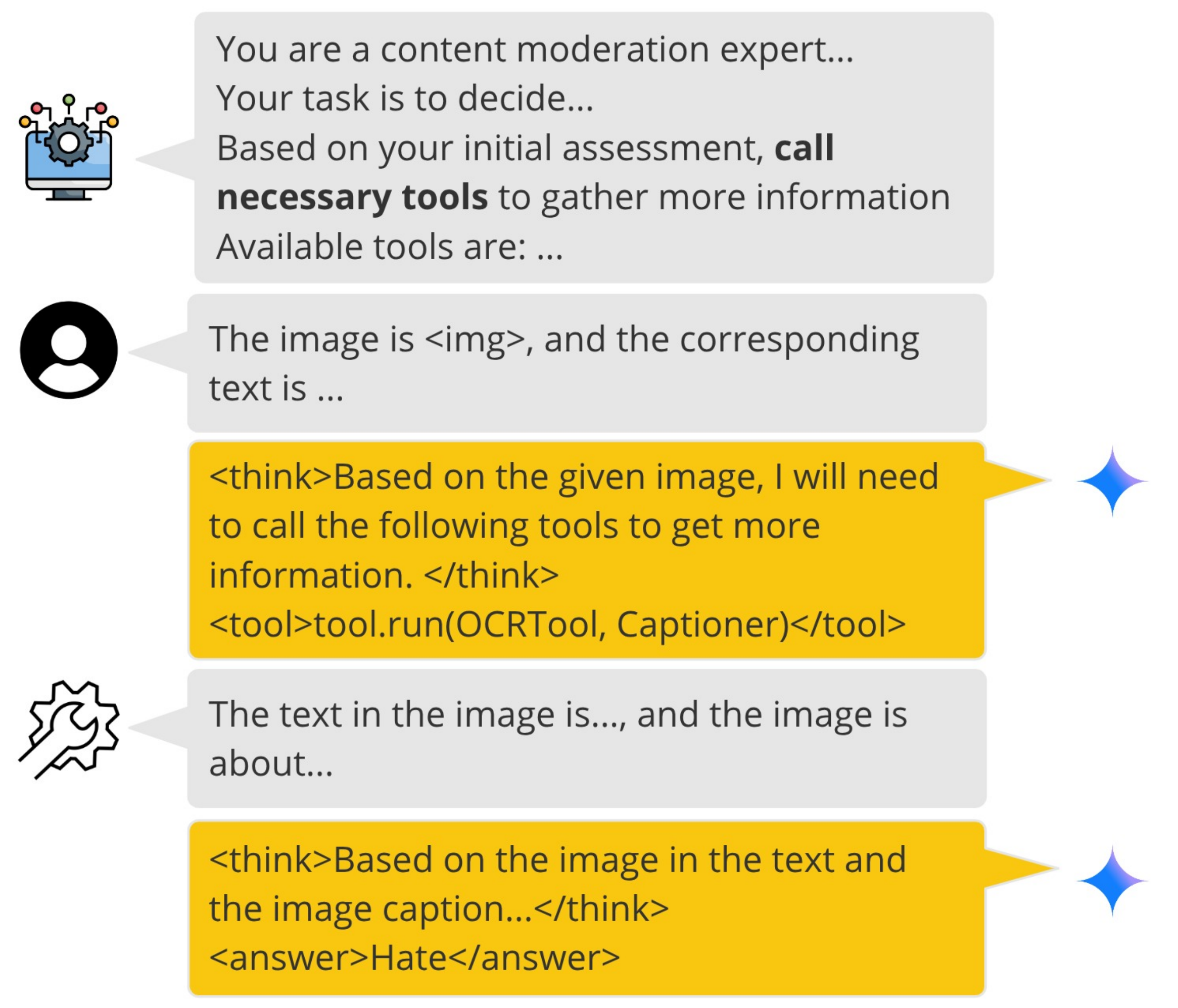} 
    \Description{Tool calling example.}
    \caption{\textbf{Multi-turn tool calling conversation.} Model select necessary tools for harder samples.}
    \label{fig:tool_calling_conversation}
\end{figure}

Figure~\ref{fig:tool_calling_conversation} illustrates a flowchart-style diagram of a multi-turn chat conversation between a system instruction, a user, an AI model, and an external tool. It models how an AI uses reasoning tags (\textit{<think>}) and function execution tags (\textit{<tool>}, \textit{<answer>}) to solve a content moderation task.

\end{document}